\pdfoutput=1

\documentclass[11pt]{article}

\usepackage[]{acl}

\usepackage{times}
\usepackage{latexsym}

\usepackage[T1]{fontenc}

\usepackage[utf8]{inputenc}

\usepackage{microtype}

\usepackage{amsmath,amssymb,amsfonts}
\usepackage{enumerate}
\usepackage{graphicx}
\usepackage{xcolor}
\usepackage{hyperref}
\usepackage{booktabs}
\usepackage{multirow,bigstrut}
\usepackage{tabu}
\usepackage{multirow, multicol}
\usepackage{adjustbox}
\usepackage{float}
\usepackage{placeins}
\usepackage{caption}

\DeclareMathOperator*{\argmin}{argmin}
\DeclareMathOperator*{\argmax}{argmax}

\newcommand{\comment}[1]{}

\title{Multimodal Conversational AI \\
A Survey of Datasets and Approaches}

\author{Anirudh S Sundar, Larry Heck \\
  Department of Electrical and Computer Engineering \\ Georgia Institute of Technology\\
  \texttt{ \{asundar34,larryheck\}@gatech.edu} \\}

\begin{document}
\maketitle

\begin{abstract}
As humans, we experience the world with all our senses or modalities (sound, sight, touch, smell, and taste). We use these modalities, particularly sight and touch, to convey and interpret specific meanings. Multimodal expressions are central to conversations; a rich set of modalities amplify and often compensate for each other. A multimodal conversational AI system answers questions, fulfills tasks, and emulates human conversations by understanding and expressing itself via multiple modalities. This paper motivates, defines, and mathematically formulates the multimodal conversational research objective. We provide a taxonomy of research required to solve the objective: multimodal representation, fusion, alignment, translation, and co-learning. We survey state-of-the-art datasets and approaches for each research area and highlight their limiting assumptions. Finally, we identify multimodal co-learning as a promising direction for multimodal conversational AI research. 
\end{abstract}

\section{Introduction}
The proliferation of smartphones has dramatically increased the frequency of interactions that humans have with digital content. These interactions have expanded over the past decade to include conversations with smartphones and in-home smart speakers. Conversational AI systems (e.g., Alexa, Siri, Google Assistant) answer questions, fulfill specific tasks, and emulate natural human conversation \cite{hakkani-tur_research_2011, gao_neural_2019}. 

Early examples of conversational AI include those based on primitive rule-based methods such as ELIZA \cite{weizenbaum_elizacomputer_1966}. More recently, conversational systems were driven by statistical machine translation systems: translating input queries to responses \cite{ritter_data-driven_2011, hakkani2012translating}. Orders of magnitude more data led to unprecedented advances in conversational technology in the mid-part of the last decade. Techniques were developed to mine conversational training data from the web search query-click stream \cite{hakkani2011exploiting, heck2012_slt, hakkani2013using} and web-based knowledge graphs \cite{heck2012exploiting, el2014extending}. With this increase in data, deep neural-networks gained momentum in conversational systems  \cite{mesnil2014using, heck2014deep, sordoni_neural_2015, vinyals_neural_2015,shang_neural_2015, serban_building_2016, li_diversity-promoting_2016, li_persona-based_2016}. 

Most recently, specialized deep learning-based conversational agents were developed primarily for three tasks: (1) goal-directed tasks in research systems \cite{shah2016interactive,eric_key-value_2017, liu2017end, liu2018dialogue, li_pumice_2019, hosseini-asl_simple_2020, wu_tod-bert_2020,peng_soloist_2021,xu_grounding_2021} and commercial products
(Siri, Cortana, Alexa, and Google Assistant), (2) question-answering \cite{yi_neural-symbolic_2019, raffel_exploring_2020, zaheer_big_2021}, and (3) open-domain conversations \cite{wolf_transfertransfo_2019,zhou_design_2020, adiwardana_towards_2020, paranjape_neural_2020, roller_open-domain_2020,bao_plato_2020,henderson_convert_2020,zhang_dialogpt_2020}. However, developing a single system with a unified approach that achieves human-level performance on all three tasks has proven elusive and is still an open problem in conversational AI.

One limitation of existing agents is that they often rely exclusively on language to communicate with users. This contrasts with humans, who converse with each other through a multitude of senses. These senses or modalities complement each other, resolving ambiguities and emphasizing ideas to make conversations meaningful. Prosody, auditory expressions of emotion, and backchannel agreement supplement speech, lip-reading disambiguates unclear words, gesticulation makes spatial references, and high-fives signify celebration. 

Alleviating this unimodal limitation of conversational AI systems requires developing methods to extract, combine, and understand information streams from multiple modalities and generate multimodal responses while simultaneously maintaining an intelligent conversation. 

Similar to the taxonomy of multimodal machine learning research \cite{baltrusaitis_multimodal_2017}, the research required to extend conversational AI systems to multiple modalities can be grouped into five areas: Representation, Fusion, Translation, Alignment, and Co-Learning. Representation and fusion involve learning mathematical constructs to mimic sensory modalities. Translation maps relationships between modalities for cross-modal reasoning. Alignment identifies regions of relevance across modalities to identify correspondences between them. Co-learning exploits the synergies across modalities by leveraging resource-rich modalities to train resource-poor modalities.

Concurrently, it is necessary for the research areas outlined above to address four main challenges in multimodal conversational reasoning -- disambiguation, response generation, coreference resolution, and dialogue state tracking \cite{kottur_simmc_2021}. Multimodal disambiguation and response generation are challenges associated with fusion that determine whether available multimodal inputs are sufficient for a direct response or if follow-up queries are required. Multimodal coreference resolution is a challenge in both translation and alignment, where the conversational agent must resolve referential mentions in dialogue to corresponding objects in other modalities. Multimodal dialogue state tracking is a holistic challenge across research areas typically associated with task-oriented systems. The goal is to parse multimodal signals to infer and update values for slots in user utterances. 

\begin{figure}[h]
    \centering
    \includegraphics[width=\columnwidth]{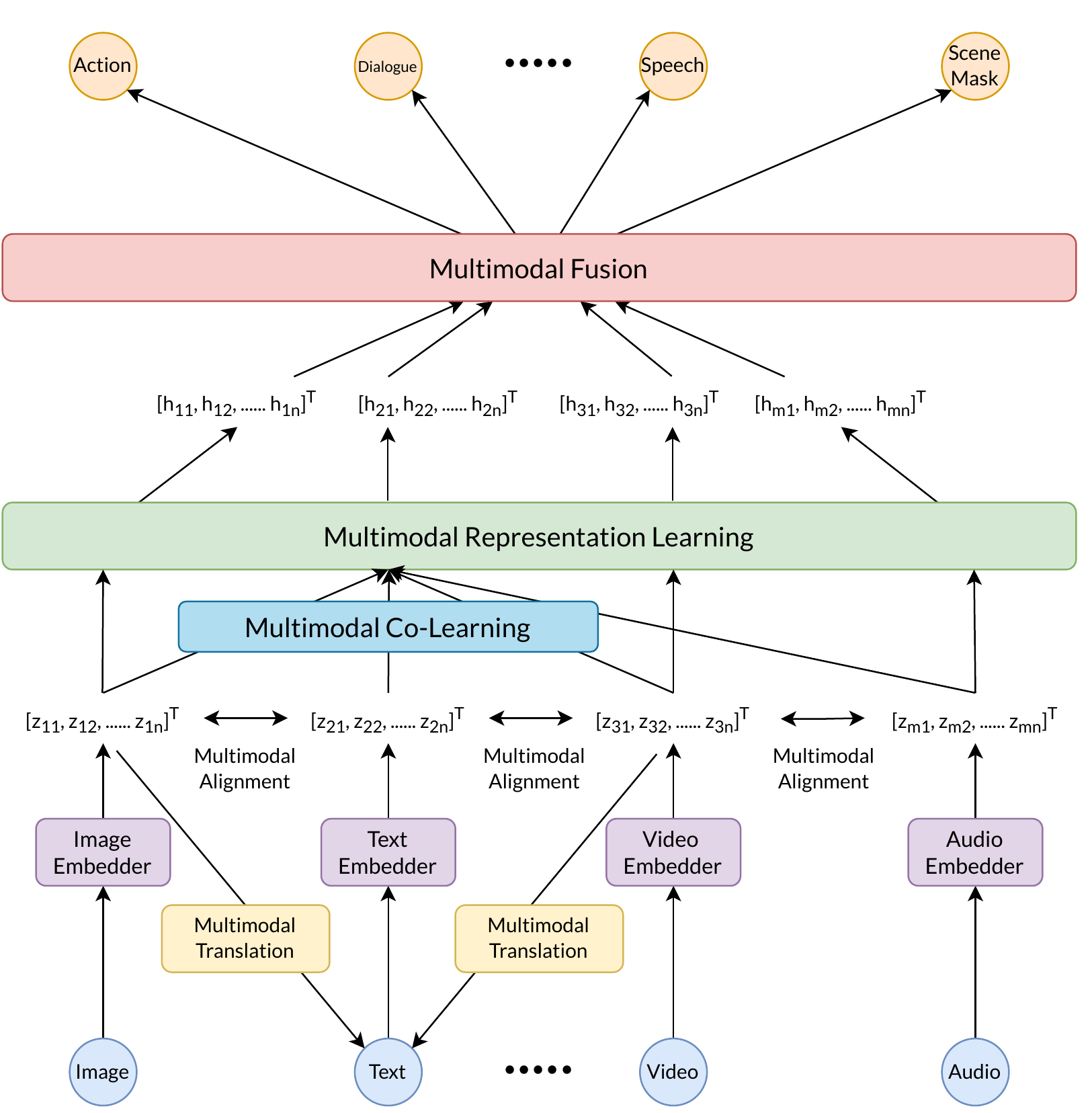}
    \caption{Taxonomy of multimodal Conversational AI research}
    \label{fig:fig_mimo}
\end{figure}

In this paper, we discuss the taxonomy of research challenges in multimodal Conversational AI as illustrated in Figure \ref{fig:fig_mimo}. Section \ref{sec:background} provides a history of research in multimodal conversations. In Section \ref{sec:math}, we  mathematically formulate multimodal conversational AI as an optimization problem. Sections \ref{sec:rep+fus}, \ref{sec:trans}, and \ref{sec:align} survey existing datasets and state-of-the-art approaches for multimodal representation and fusion, translation, and alignment. Section \ref{sec:disc} highlights limitations of existing research in multimodal conversational AI and explores multimodal co-learning as a promising direction for research. 

\section{Background} \label{sec:background}
Early work in multimodal conversational AI focused on the use of visual information to improve automatic speech recognition (ASR). One of the earliest papers along these lines is by \citet{yuhas_integration_1989} followed by many papers including work by \citet{meier_adaptive_1996}, \citet{ duchnowski_see_1994}, \citet{ bregler_eigenlips_1994}, and \citet{ngiam_multimodal_2011}. 

Advances in client-side capabilities enabled ASR systems to utilize other modalities such as tactile, voice, and text inputs. These systems supported more comprehensive interactions and facilitated a higher degree of personalization. Examples include \textsc{Esprit}'s MASK \cite{lamel_user_1998}, Microsoft's MiPad \cite{huang_mipad_2001}, and AT\&T's MATCH \cite{johnston_match_2002}.

Vision-driven tasks motivated research in adding visual understanding technology into conversational AI systems.
Early work in reasoning over text+video include work by \citet{ramanathan_linking_2014} where they leveraged these combined modalities to address the problem of assigning names
of people in the cast to tracks in TV videos.  \citet{kong_what_2014} leveraged natural language descrptions of RGB-D videos for 3D semantic parsing. \citet{srivastava_multimodal_2014} developed a multimodal Deep Boltzmann Machine for image-text retrieval and ASR using videos. \citet{antol_vqa_2015} introduced a dataset and baselines for multimodal question-answering, a challenge combining computer vision and natural language processing. More recent work by  \citet{zhang2019generative} and \citet{selvaraju_taking_2019} leveraged conversational explanations to make vision and language models more grounded, resulting in improved visual question answering.

While modalities most commonly considered in the conversational AI literature are text, vision, tactile, and speech, other sources of information are gaining popularity within the research community. These include eye-gaze, 3D scans, emotion, action and dialogue history, and virtual reality. \citet{heck_multimodal_2013} and \citet{hakkani-tur_eye_2014} use gesture, speech, and eye-gaze to resolve and infer intent in conversational web-browsing systems. \citet{grauman_ego4d_2021} presents ego-centric video understanding, \citet{padmakumar_teach_2021}, and \citet{shridhar_alfred_2020} present task completion from 3D simulations, and \citet{gao_objectfolder_2021} presents multisensory object recognition. 

Processing conventional and new modalities brings forth numerous challenges for multimodal conversations. To answer these challenges, we will first mathematically formulate the multimodal conversational AI problem, then detail fundamental research sub-tasks required to solve it.

\section{Mathematical Formulation} \label{sec:math} 
We formulate multimodal conversational AI 
as an optimization problem. The objective is to find the optimal response $\mathbf{S}$ to a message $m$ given underlying multimodal context~$c$. Based on the sufficiency of the context, the optimal response could be a statement of fact or a follow-up question to resolve ambiguities. Statistically, $\mathbf{S}$ is estimated as:

\begin{equation}
    \mathbf{S} = \argmax_{r} p(r \vert c,m).
    \label{eq:conv_AI}
\end{equation}

The probability of an arbitrary response $r$ can be expressed as a product of the probabilities of responses $\{r_i\}_{i=1}^T$ over $T$ turns of conversation  \cite{sordoni_neural_2015}. 
\begin{equation}
    p(r \vert c,m) = \prod_{i=1}^{T}p(r_i \vert r_1, \hdots r_{i-1}, c,m)
    \label{eq:prob_exp}
\end{equation}



It is also possible for conversational AI to respond through multiple modalities. We represent the multimodality of output responses by a matrix $R:=\{ r_i^1, r_i^2, \hdots r_i^l \}$ over $l$ permissible output modalities. 
\begin{equation}
    \mathbf{S} = \argmax_{R} p(R \vert c,m)
    \label{eq:mm_output}
\end{equation}

Learning from multimodal data requires manipulating information from all modalities using a function $f(\cdot)$ consisting of five sub-tasks: representation, fusion, translation, alignment, and co-learning.  We include these modifications and present the final multimodal conversational objective below.

\begin{equation}
    \mathbf{S} = \argmax_{R} p(R \vert f(c,m)) 
    \label{eq:max}
\end{equation}

In the following sections, we describe each sub-task contained in $f(\cdot)$. 



\section{Multimodal Representation + Fusion} \label{sec:rep+fus}
Multimodal representation learning and fusion are primary challenges in multimodal conversations.
Multimodal representation is the encoding of multimodal data in a format amenable to computational processing. Multimodal fusion concerns joining features from multiple modalities to make predictions.

\subsection{Multimodal Representations}
Using multimodal information of varying granularity for conversations necessitates techniques to represent high-dimensional signals in a latent space. These latent multimodal representations encode human senses to improve a conversational AI's perception of the real-world. Success in multimodal tasks requires that representations satisfy three desiderata \cite{srivastava_multimodal_2014}:
\begin{enumerate}
    \item 
    Similarity in the representation space implies similarity of the corresponding concepts
    \item 
    The representation is easy to obtain in the absence of some modalities
    \item 
    It is possible to infer missing information from observed modalities
\end{enumerate}

There exist numerous representation methods for the range of problems multimodal conversational AI addresses.
Multimodal representations are broadly classified as either joint representations or coordinated representations~\cite{baltrusaitis_multimodal_2017}.

\subsubsection{Joint Representations}
Joint representations combine unimodal signals into the same representation space. Traditional techniques to learn joint representations include multimodal autoencoders \cite{ngiam_multimodal_2011}, multimodal deep belief networks \cite{srivastava_multimodal_2014}, and sequential networks \cite{nicolaou_continuous_2011}. 

The success of the Transformer to represent text \cite{vaswani_attention_2017} and BERT when modeling language \cite{devlin_bert_2019} have inspired a variety of multimodal transformer-based architectures for (1) vision-and-language understanding \cite{sun_videobert_2019, lu_vilbert_2019, gabeur_multi-modal_2020, chen_uniter_2020, tan2019lxmert,singh_flava_2021}, (2) vision-grounded speech recognition \cite{baevski_wav2vec_2020, hsu_hubert_2021, chan_multi-modal_2021}, and (3) User Interface (UI) understanding \cite{bapna2017towards, he_actionbert_2021,bai_uibert_2021,li_vut_2021,xu_grounding_2021, heck_zero-shot_2022}. 

Transformer-based models used as joint multimodal representations can be described as illustrated in the taxonomy of Figure \ref{fig:fig_mimo}. Modality specific encoders $\{j_i(\cdot)\}_{i=1}^n$ embed unimodal tokens $\{c_{i_{k}}\}_{k=1}^n$ to create latent features $\{z_{i_{k}}\}_{k=1}^n$ (Equation \ref{eq:j_mm_1}). Decoder networks use latent features to produce output symbols. A transformer $\Psi(\cdot)$ consists of stacked encoders and decoders with intra-modality attention. Attention heads compute relationships within elements of a modality, producing multimodal representations~$\{h_{i_{k}}\}_{k=1}^n$ (Equation \ref{eq:j_mm_2}).

\begin{gather}
    z_{i_1}, z_{i_2}, \hdots z_{i_n} =  j_i(c_{i_1}, c_{i_2}, \hdots c_{i_n})
    \label{eq:j_mm_1} \\
    h_{1_1} \hdots h_{{m_n}} = \Psi(z_{1_1}, z_{1_2}, \hdots z_{m_n}) \label{eq:j_mm_2}
\end{gather}



\subsubsection{Coordinated Representations}
In contrast, coordinated representations model each modality separately. Constraints coordinate representations of separate modalities by enforcing cross-modal similarity over concepts. For example, the audio representation $g_a(\cdot)$ of a dog's bark would be closer to the dog's image representation $g_i(\cdot)$ and further away from a car's (Equation \ref{eq:coord_rep}). A notion of distance $d$ between modalities in the coordinated space enables cross-modal retrieval.
\begin{equation}
    d(g_{a}(\text{dog}), g_{i}(\text{dog})) < d(g_{a}(\text{dog}), g_{i}(\text{car}))
    \label{eq:coord_rep}
\end{equation}

In practice, contrastive objectives are used to coordinate representations between pairs of modalities. Contrastive learning has been successful in relating separate views of the same image \cite{becker_self-organizing_1992,chen_simple_2020, he_momentum_2020,grill_bootstrap_2020,radford_learning_2021-1}, images and their natural language descriptions \cite{weston_large_2010,kiros_unifying_2014,zhang_contrastive_2020,li_oscar_2020}, and videos with their corresponding audio and natural language descriptions \cite{owens_ambient_2016,korbar_cooperative_2018,sun_learning_2019,  miech_end--end_2020, alayrac_self-supervised_2020,akbari_vatt_2021, xu_videoclip_2021,qian_spatiotemporal_2021,morgado_audio-visual_2021}.

\subsection{Multimodal Fusion}
Multimodal fusion combines features from multiple modalities to make decisions, denoted by the final block before the outputs in Figure \ref{fig:fig_mimo}. Fusion approaches are broadly classified into model-agnostic and model-based methods.

Model-agnostic methods are independent of specific algorithms and are split into early, late, and hybrid fusion. Early fusion integrates features following extraction, projecting features into a shared space \cite{potamianos_recent_2003,ngiam_multimodal_2011, nicolaou_continuous_2011,jansen_coincidence_2019}. In contrast, late fusion integrates decisions from unimodal predictors \cite{becker_self-organizing_1992,korbar_cooperative_2018,shuster_image_2020, alayrac_self-supervised_2020, akbari_vatt_2021}. Early fusion is predominantly used to combine features extracted in joint representations while late fusion combines decisions made in coordinated representations. Hybrid fusion exploits both low and high level modality interactions \cite{hutchison_multi-level_2005-1,schwartz_factor_2020, piergiovanni_evolving_2020,goyal_cross-modal_2020-1}.

Model-based methods consist of graphical techniques like Hidden Markov Models \cite{nefian_coupled_2002, gurban_dynamic_2008}, neural networks \cite{nicolaou_continuous_2011, antol_vqa_2015,gao_are_2015,malinowski_ask_2015,kottur_visual_2018,qian_spatiotemporal_2021}, and transformers \cite{xu_ask_2016, hori_attention-based_2017,peng_dynamic_2019,zhang_generative_2019,shuster_image_2020, chen_uniter_2020,geng_dynamic_2021,xu_grounding_2021} 

\subsection{State-of-the-art Representation+Fusion Models for Conversational AI}
Having introduced the multimodal representation and fusion challenges, we present the state-of-the-art in these sub-tasks for conversational AI. 

\subsubsection{Factor Graph Attention}
\citet{schwartz_factor_2020} develops Factor Graph Attention (FGA), a joint representation for multi-turn question answering grounded in images.
FGA embeds images using VGG-16 \cite{simonyan_very_2015} or F-RCNN \cite{ren_faster_2016} and textual modalities using LSTMs. 
Nodes in the factor graph represent attention distributions over elements of each modality, and factors capture relationships between nodes.

There are two types of factors -- local and joint. Local factors capture interactions between nodes of a single modality (e.g., words in the same sentence), while joint factors capture interactions between different modalities (e.g., a word in a sentence and an object in an image). 

Representations from all modalities are concatenated via hybrid fusion and passed through a multi-layer perceptron network to retrieve the best candidate answer.

Table \ref{tab:fga_accuracy} compares the Recall-at-k (R@k) of discriminative models on VisDial v1.0 test-std. The F-RCNN version of FGA is the state-of-the-art.  

\begin{table}[ht]
    \renewcommand{\arraystretch}{1}
    \centering
    \begin{adjustbox}{width=\columnwidth,center}
    \begin{tabu}{c|c|c|c}
        \toprule
         Model & R@1 & R@5 & R@10 \\
         \hline 
         LF \cite{das_visual_2017} &  40.95 & 72.45 & 82.83\\
         HRE \cite{das_visual_2017} &  39.93 & 70.45 & 81.50 \\
         Memory Network \cite{das_visual_2017} & 40.98 & 72.30 & 83.30  \\
         CorefNMN (ResNet-152) \cite{kottur_visual_2018} & 47.55 & 78.10 & 88.80 \\
         NMN (ResNet-152) \cite{hu_learning_2017}  & 44.15 & 76.88 & 86.88 \\
         \textbf{FGA (F-RCNNx101)} \cite{schwartz_factor_2020} & \textbf{52.75} & \textbf{82.92} & \textbf{91.07} \\
         \bottomrule
    \end{tabu}
    \end{adjustbox}
    \caption{Comparison of models on VisDial v1.0 test-std (Recall@k)\cite{schwartz_factor_2020}}
    \label{tab:fga_accuracy}
\end{table}

\subsubsection{\textsc{TransResNet}}
\citet{shuster_image_2020} presents \textsc{TransResNet} for image-based dialogue. Image-based dialogue is the task of choosing the optimal response on a dialogue turn given an image, an agent personality, and dialogue history. \textsc{TransResNet} consists of separately learned sub-networks to represent input modalities. Images are encoded using ResNeXt 32×48d trained on 3.5 billion Instagram images \cite{xie_aggregated_2017}, personalities are embedded using a linear layer, and dialogue is encoded by a transformer pretrained on Reddit \cite{mazare_training_2018} to create a joint representation. 

\textsc{TransResNet} compares model-agnostic and model-based fusion by using either concatenation or attention networks to combine representations. Like FGA, the chosen dialogue response is the candidate closest to the fused representation.

On the first turn, \textsc{TransResNet} uses only style and image information to produce responses. Dialogue history serves as an additional modality on subsequent rounds. Ablation of one or more modalities diminishes the ability of the model to retrieve the correct response. Optimal performance on Image-Chat \cite{shuster_image_2020} is achieved using multimodal concatenation of jointly represented modalities (Table \ref{tab:transresnet_accuracy}).
\begin{table}[ht]
    \renewcommand{\arraystretch}{1}
    \centering
    \begin{adjustbox}{width=\columnwidth,center}
    \begin{tabu}{c|c|c|c|c}
        \toprule
         Modalities & Turn 1 & Turn 2 & Turn 3 & All \\
         \hline 
         Image Only & 37.6 & 28.1 & 20.7 & 28.7 \\
         Style Only & 18.3 & 15.3 & 17.0 & 16.9 \\
         Dialogue History Only & 1.0 & 33.7 & 32.3 & 22.3 \\
         \hline 
         Style + Dialogue & 18.3 & 45.4 & 43.1 & 35.4 \\
         Image + Dialogue & 37.6 & 39.4 & 32.6 & 36.5 \\
         Image + Style & \textbf{54.0} & 41.1 & 35.2 & 43.4 \\
         \hline 
         Style + Dialogue + Image & \textbf{54.0} & \textbf{51.9} & \textbf{44.8} & \textbf{50.3} \\
         \bottomrule
    \end{tabu}
    \end{adjustbox}
    \caption{Recall@1 (\%) on Image-Chat using \textsc{TransResNet\textsubscript{RET}} (ResNeXt-IG-3.5B, MM-Sum)}
    \label{tab:transresnet_accuracy}
\end{table}

\subsubsection{MultiModal Versatile Networks (MMV)}
\citet{alayrac_self-supervised_2020}  presents a training strategy to learn coordinated representations using self-supervised contrastive learning from instructional videos. Videos are encoded using TSM with a ResNet50 backbone \cite{lin_tsm_2019}, audio is encoded using log MEL spectrograms from ResNet50, and text is encoded using Google News pre-trained word2vec \cite{mikolov_efficient_2013}.

\citet{alayrac_self-supervised_2020} defines three types of coordinated spaces: shared, disjoint, and `fine+coarse'.
The shared space enables direct comparison and navigation between modalities, by assuming equal granularity. The disjoint space sidesteps navigation to solve the granularity problem by creating a space for each pair of modalities. The `fine+coarse' space solves both issues by learning two spaces. A fine-grained space compares audio and video, while a lower-dimensional coarse-grained space compares fine-grained embeddings with text. We further discuss the MMV model in Section \ref{subsec:mmv}.


\section{Multimodal Translation} \label{sec:trans}
Multimodal translation maps embeddings from one modality to signals from another for cross-modality reasoning (Figure \ref{fig:fig_mimo}).  Cross-modal reasoning enables multimodal conversational AI to hold meaningful conversations and resolve references across multiple senses, specifically language and vision. To this end, we survey existing work addressing the translation of images and videos to text. We discuss multimodal question-answering and multimodal dialogue, translation tasks that extend to multimodal conversations.

\subsection{Image}
\citet{antol_vqa_2015} and \citet{zhu_visual7w_2016} present Visual Question-Answering (VQA) and Visual7W for multimodal question answering (MQA). The MQA challenge requires responding to textual queries about an image. Both datasets collect questions and answers using crowd workers, encouraging trained models to learn natural responses. \citet{heck_zero-shot_2022} presents the Visual Slot dataset, where trained models learn answers to questions grounded in UIs. 

The objective of MQA is a simplification of Equation \ref{eq:max} to a single-turn, single-timestep scenario ($T=1$), producing a response to a question $m_q$ given multimodal context $\{c_{i}\}_{i=1}^{n}$:
\begin{equation}
    \mathbf{S}_{MQA} = \argmax_{R} p(R \vert  f(c_1, \hdots c_{n},m_q))
    \label{eq:obj_QA}
\end{equation}

Multi-turn question-answering (MTQA) is the next step towards multimodal conversational AI. VisDial  \cite{das_visual_2017} extends VQA to multiple turns, translating over QA history in addition to images. 
GuessWhat?! \cite{de_vries_guesswhat_2017} is a guessing game, discovering objects in a scene through dialogue.  \textsc{ManyModal}QA \cite{hannan_manymodalqa_2020} requires reasoning over prior knowledge, images, and databases. MIMOQA \cite{singh_mimoqa_2021} is an example of multimodal responses, where answers are image-text pairs. 

The objective of MTQA (Equation \ref{eq:obj_mtqa}) is an extension of MQA to include QA history $\mathbf{h}_{qa}~=~\{m_{q_{1}}, r_{a_{1}},m_{q_{2}}, r_{a_{2}}, \hdots m_{q_{i-1}},r_{a_{i-1}} \}$. 
\begin{equation}
    \mathbf{S}_{MTQA} = \argmax_R p(R \vert  f(c_1,\hdots c_n,\mathbf{h}_{qa}))
    \label{eq:obj_mtqa}
\end{equation}

Image-Grounded Conversations (IGC) \cite{mostafazadeh_image-grounded_2017} builds on MTQA by presenting a dataset for multimodal dialogue~(MD): machine perception and conversation through language. Image-Chat \cite{shuster_image_2020} extends IGC to agents with personalities. Crowd workers hold three-turn conversations about an image with one of 215 emotions (e.g., peaceful, erratic, skeptical). Motivated by the popularity of visual content in instant-messaging, Meme
incorporated Open-domain Dialogue (MOD) \cite{fei_towards_2021} contains natural language conversations interspersed with behavioral stickers. SIMMC \cite{moon_situated_2020} and SIMMC2.0 \cite{kottur_simmc_2021} present goal-oriented dialogue for shopping. The challenge requires leveraging dialogue and a state of the world to resolve references, track dialogue state, and recommend the correct object. IGC, Image-Chat, MOD, SIMMC, and SIMMC2.0 solve the MD objective that depends on previous dialogue responses $\mathbf{h}_d=\{m_{d_1},r_{d_1},m_{d_2},r_{d_2},\hdots m_{d_{i-1}},r_{d_{i-1}} \}$:
\begin{equation}
    \mathbf{S}_{MD} = \argmax_R p(R \vert  f(c_1, \hdots c_{n},\mathbf{h}_{d}))
    \label{eq:obj_mtd}
\end{equation}


\subsection{Video}
An extension of VQA to the video domain includes TVQA, TVQA+ \cite{lei_tvqa_2020} built on TV shows, MovieQA \cite{tapaswi_movieqa_2016} based on movies, and Audio Visual Scene-Aware Dialog (AVSD) \cite{alamri_audio-visual_2019} based on CHARADES \cite{sigurdsson_hollywood_2016}. DVD \cite{le_dvd_2021} presents video-QA over videos synthesized from the CATER dataset \cite{Girdhar2020CATER:}.
Besides visual reasoning, video-QA requires temporal reasoning, a challenge addressed by multimodal alignment that we discuss in the following section. 

\section{Multimodal Alignment} \label{sec:align}
While image-based dialogue revolves around objects (e.g., cats and dogs), video-based dialogue revolves around objects and associated actions (e.g., jumping cats and barking dogs) where spatial and temporal features serve as building blocks for conversations. Extracting these spatiotemporal features requires multimodal alignment -- aligning sub-components of different modalities to find correspondences. We identify action recognition and action from modalities as alignment challenges relevant to multimodal conversations. 

\subsection{Action Recognition}
Action recognition is the task of extracting natural language descriptions from videos. 
UCF101 \cite{soomro_ucf101_2012}, HMDB51 \cite{kuehne_hmdb_2011}, and Kinetics-700 \cite{carreira_short_2019} involve extracting actions from short YouTube and Hollywood movie clips. HowTo100M \cite{miech_howto100m_2019}, MSR-VTT \cite{xu_msr-vtt_2016}, and YouCook2 \cite{zhou_towards_2017} are datasets containing instructional videos on the internet and require learning text-video embeddings. YouCook2 and MSR-VTT are annotated by hand while HowTo100M uses existing video subtitles or ASR. 

Mathematically, the goal is to retrieve the correct natural language description $\mathbf{y} \in \mathcal{Y}$ to a query video $\mathbf{x}$ (Equation \ref{eq:action_rec}). Video and text representation functions $g(\cdot)_\text{video}$ and $g(\cdot)_\text{text}$ embed modalities into a coordinated space where they are compared using a distance measure $d$.
\begin{equation}
    \argmin_{y \in \mathcal{Y}} d(g_\text{video}(\mathbf{x}), g_\text{text}(\mathbf{y}_j))
    \label{eq:action_rec}
\end{equation}


\subsection{Action from Modalities}
Equipping multimodal conversational agents with the ability to perform actions from multiple modalities provides them with an understanding of the real world, improving their conversational utility. 

Talk the Walk \cite{de_vries_talk_2018} presents the task of navigation conditioned on partial information. A ``tourist'' provides descriptions of a photo-realistic environment to a ``guide'' who determines actions.  
Vision-and-Dialog Navigation \cite{thomason_vision-and-dialog_2019} contains natural dialogues grounded in a simulated environment. The task is to predict a sequence of actions to a goal state given the world scene, dialogue, and previous actions.
TEACh \cite{padmakumar_teach_2021} extends Vision-and-Dialog Navigation to complete tasks in an AI2-THOR simulation. The challenge involves aligning information from language, video, as well as action and dialogue history to solve daily tasks.
Ego4D \cite{grauman_ego4d_2021} contains text annotated egocentric (first person) videos in real-world scenarios. Ego4D includes 3D scans, multiple camera views, and eye gaze, presenting new representation, fusion, translation, and alignment challenges. It is associated with five benchmarks: Video QA, object state tracking, audio-visual diarization, social cue detection, and camera trajectory forecasting. 

\subsection{Multimodal Versatile Networks (MMV)} \label{subsec:mmv}

\begin{table*}[h]
    \renewcommand{\arraystretch}{1}
    \centering
    \begin{adjustbox}{width=\textwidth,center}
    \begin{tabu}{c|c|c|c|c|c|c|c}
        \toprule
          Model & UCF101 (FT) & HMDB51 (FT) & ESC-50 (Linear) & AS & K600 & YC2 & MSR-VTT\\ 
         \hline 
         MIL-NCE (S3D-G) \cite{miech_end--end_2020} & 91.3 & 61.0 & / & / & / & \textbf{51.2} & \textbf{32.4} \\
         AVTS (MC3) \cite{korbar_cooperative_2018}  & 89.0 & 61.6 & 80.6 & / & / & / & / \\
         AA+AV CC \cite{jansen_coincidence_2019} & / & / & / & 28.5 & /& / & / \\
         CVRL \cite{qian_spatiotemporal_2021} & / & / & / & / & 64.1 & / & / \\
         XDC \cite{alwassel_self-supervised_2020} & 91.2 & 61.0 & 84.8 & / & / & / & / \\
         ELo \cite{piergiovanni_evolving_2020} & 93.8 & 67.4 & / & / & / & / & / \\
         AVID \cite{morgado_audio-visual_2021} & 91.5 & 64.7 & \textbf{89.2} & / & / & / & / \\
         GDT (IG65M) \cite{patrick_compositions_nodate} & \textbf{95.2} & 72.8 & 88.5 & / & / & / & / \\
         MMV FAC (TSM-50x2) \cite{alayrac_self-supervised_2020} & \textbf{95.2} & \textbf{75.0} & 88.9 & \textbf{30.9} & \textbf{70.5} & 45.4& 31.1\\ 
         
         \bottomrule
    \end{tabu}
    \end{adjustbox}
    \caption{Comparison of learnt representations on UCF101, HMDB51, ESC-50, \underline{A}udio\underline{S}et, \underline{K}inetics\underline{600}, \underline{Y}ou\underline{C}ook\underline{2}, and MSR-VTT. Top-1 Accuracy for UCF101, HMDB51, ESC-50, Kinetics600, mean Average Precision (mAP) for AudioSet, Recall@10 for YouCook2 and MSR-VTT \cite{alayrac_self-supervised_2020}. }
    \label{tab:mmv_accuracy}
\end{table*}

In addition to a representation, \citet{alayrac_self-supervised_2020} presents a self-supervised task to train modality embedding graphs for multimodal alignment. Sampling temporally aligned audio, visual clips, and narrations from the same video creates positive training examples, while those from different videos comprise negative training examples. A Noise-Contrastive Estimation (NCE) loss \cite{pmlr-v9-gutmann10a} is minimized to ensure similarity between embeddings of positive training examples while forcing negative pairs further apart. A Multiple Instance Learning (MIL) \cite{miech_end--end_2020} variant of NCE measures loss on pairs of modalities of different granularity. MIL accounts for misalignment between audio/video and text by measuring the loss of fine-grained information with multiple temporally close narrations.

The network is trained on HowTo100M \cite{miech_howto100m_2019} and AudioSet \cite{gemmeke_audio_2017}. Table \ref{tab:mmv_accuracy} compares the performance of MMV on action classification, audio classification, and zero-shot text-to-video retrieval.

\section{Discussion} \label{sec:disc}
The current datasets used for research in multimodal conversational AI are summarized in Table~\ref{tab:dataset_summary}. While MQA and MTQA are promising starting points for multimodal natural language tasks, extending QA to conversations is not straightforward. Inherently, MQA limits itself to direct questions targeting visible content, whereas multimodal conversations require understanding information that is often implied \cite{mostafazadeh_image-grounded_2017}. Utterances in dialogue represent speech acts and are classified as constatives, directives, commissives, or acknowledgments \cite{bach_linguistic_1979}. Answers belong to a single speech act (constatives) and represent a subset of natural conversations.

Similarly, the work to-date on action recognition is incomplete and insufficient for conversational systems. Conversational AI must represent and understand spatiotemporal interactions. However, current research in action recognition attempts to learn relationships between videos and their natural language descriptions. These descriptions are not speech acts themselves. Therefore, they do not adequately represent dialogue but rather only serve as anchor points in the interaction. 

\begin{table*}[t]
  \centering
  
  \begin{adjustbox}{width=0.99\textwidth,center}
  \begin{tabular}{|l|l|l|l|l|}
  \hline
    Dataset & Modalities & Task & Data Collection & POV\\
    \hline 
    VQA \cite{antol_vqa_2015} & I,Q & Question Answering & Human-Human & Third Person \\
    Visual7W \cite{zhu_visual7w_2016} & I,Q & Question Answering & Human-Human & Third Person \\
    Visual Slot \cite{heck_zero-shot_2022} & UI, Q & Question Answering & Human & X \\
    TVQA \cite{lei_tvqa_2019} & V,Q,S & Question Answering & Human-Human & Third Person \\
    MovieQA \cite{tapaswi_movieqa_2016} & V,C,Q,T,S & Question Answering & Human & Third Person \\
    \textsc{ManyModal}QA \cite{hannan_manymodalqa_2020} & I,C,Q,T,Tables & Question Answering & Human-Human & Third Person\\
    MIMOQA \cite{singh_mimoqa_2021} & I,Q,T & Question Answering & Machine & X \\
    VisDial \cite{das_visual_2017} & I,H\textsubscript{Q},H\textsubscript{A},C,Q & Question Answering & Human-Human & Third person \\
    Guesswhat \cite{de_vries_guesswhat_2017} & I, H\textsubscript{Q},H\textsubscript{A} & Question Answering &Human-Human & Third Person \\
    AVSD \cite{alamri_audio-visual_2019} & V,A,H\textsubscript{Q},H\textsubscript{A},C,Q & Question Answering & Human-Human & Third Person \\
    DVD \cite{le_dvd_2021} & V,Q,H\textsubscript{Q},H\textsubscript{A}  & Question Answering & Machine & X \\
    SIMMC \cite{moon_situated_2020} & H\textsubscript{D}, Q, VR  &  Shopping & Machine Self-play & First Person \\
    SIMMC2.0 \cite{kottur_simmc_2021} & H\textsubscript{D}, Q, VR  &  Shopping & Machine Self-play & First Person \\
    IGC \cite{mostafazadeh_image-grounded_2017} & I,Q,D & Chit-chat + Question Answering & Human-Human & Third-Person\\
    Image-Chat \cite{shuster_image_2020} & I,D,Personality & Chit-chat & Human-Human & Third Person \\
    MOD \cite{fei_towards_2021} & D, Personality & Visual chit-chat & Human-Human & X \\
    UCF101 \cite{soomro_ucf101_2012} & V,A (partial) & Action Recognition & YouTube & Third Person \\
    HMDB51 \cite{kuehne_hmdb_2011} & V & Action Recognition & YouTube+Movies & Third Person \\
    Kinetics 700 \cite{carreira_short_2019} & V & Action Recognition & YouTube & Third Person \\
    HowTo100M \cite{miech_howto100m_2019} & V,T,S & Text-Video Embeddings  & YouTube &  First+Third Person \\
    YouCook2 \cite{zhou_towards_2017} &V,T & Text-video retrieval, activity recognition & YouTube & Third Person \\
    MSRVTT \cite{xu_msr-vtt_2016} &V,A,T & Video-to-text & Web videos  & First+Third Person \\
    Talk the Walk \cite{de_vries_talk_2018} & I, Actions, D & Navigation from Actions and Dialogue & Human-Human & First Person \\
    CVDN \cite{thomason_vision-and-dialog_2019} & VR, Actions, H\textsubscript{Q},H\textsubscript{A}   & Navigation from  Dialogue History & Human-Human & First Person \\
    TEACh \cite{padmakumar_teach_2021} & Scene, Actions, D & Action prediction, Task from language & AI2-THOR & First+Third Person\\
    Ego4D \cite{grauman_ego4d_2021} &V,T,A,Gaze,3D Scan,S& Spatial Reasoning &Human & First Person\\
    \hline
  \end{tabular}
  \end{adjustbox}
  \caption{Datasets for multimodal representations. I=Image, V=Video, UI=User Interface, C=Caption, Q=Question, T=Text, H\textsubscript{Q}~=~Question history, H\textsubscript{A}=Answer history, H\textsubscript{D}=Dialogue history, VR=Virtual Reality, D=Dialogue, A=Audio, S=Speech}
  \label{tab:dataset_summary}
  \vspace{5 pt}

  
\end{table*}


%

In contrast, Image-Chat \cite{shuster_image_2020} presents a learning challenge directly aligned with the multimodal dialogue objective in Equation \ref{eq:max}. Image-Chat treats dialogue as an open-ended discussion grounded in the visual modality. Succeeding in the task requires jointly optimizing visual and conversational performance. The use of crowd workers that adopt personalities during data collection encourages natural dialogue and captures conversational intricacies and implicatures. 

MQA answers explicit questions about an image (\includegraphics[]{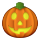}$\rightarrow$\textit{Is this at a farm?}), and action recognition describes videos (\includegraphics[]{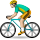}$\rightarrow$\textit{Mountain biking}). On the other hand, Image-Chat requires both implicit knowledge (\includegraphics[]{emoji_images/hires/1F383.pdf} ~$\rightarrow$~\textit{Halloween}, \includegraphics[]{emoji_images/hires/1F6B4.pdf} $\rightarrow$ \textit{Exercise}) and multi-turn reasoning (\includegraphics[]{emoji_images/hires/1F383.pdf} $\rightarrow$ \textit{Halloween} $\rightarrow$ \textit{Holiday}, \includegraphics[]{emoji_images/hires/1F6B4.pdf} $\rightarrow$ \textit{Exercise} $\rightarrow$ \textit{Fitness}). 

Despite its advantages over other datasets, Image-Chat makes three assumptions about multimodal conversations limiting its extension to the multimodal conversational objective:
\begin{enumerate}
    \item Conversations are limited to three turns, devoid of long-term dialogue dependencies.
    \item Language and images are the only modalities.
    \item Personalities are independent of previous responses. This differs from natural human conversations where humans tend to understand and reciprocate the personality of a dialogue partner \cite{rashkin_towards_2019}.
\end{enumerate}

The discussion above highlights the limitations of existing datasets for the multimodal conversational AI task. Datasets need to be improved to better capture and represent more natural, multi-turn dialogues over multiple modalities; dialogues that more closely resemble how humans converse with each other and their environment.

In addition, algorithmic improvements are required to advance the field of multimodal conversational AI - particularly with respect to the objective function. Current approaches such as MQA and action recognition models optimize a limited objective compared to Equation~\ref{eq:max}. We postulate that the degradation of these methods when applied to multimodal conversations is largely caused by this and, therefore, motivates investigation. 

Another open research problem is to improve performance on Image-Chat. The current state-of-the-art \textsc{TransResNet\textsubscript{Ret}} is limited. 
The model often hallucinates, referring to content missing in the image and previous dialogue turns. The model also struggles when answering questions and holding extended conversations. We suspect these problems are a reflection of the limiting assumptions Image-Chat makes and the absence of multimodal co-learning to extract relationships between modalities. 
For further details, we refer readers to example conversations in Appendix \ref{sec:appendix_transresnet}. 

Different modalities often contain complementary information when grounded in the same concept. Multimodal co-learning exploits this cross-modality synergy to model resource-poor modalities using resource-rich modalities. An example of co-learning in context of Figure \ref{fig:fig_mimo} is the use of visual information and audio to generate contextualized text representations. 

\citet{blum_combining_1998} introduced an early approach to multimodal co-training, using information from hyperlinked pages for web-page classification. \citet{socher_connecting_2010} and \citet{duan_multimodal_2014} presented weakly-supervised techniques to tag images given information from other modalities. \citet{kiela_grounding_2015} grounded natural language descriptions in olfactory data. More recently, \citet{upadhyay_almost_2018} jointly trains bilingual models to accelerate spoken language understanding in low resource languages. \citet{selvaraju_taking_2019} uses human attention maps to teach QA agents ``where to look''. Despite the rich history of work in multimodal co-learning, extending these techniques to develop multimodal conversational AI that understands and leverages cross-modal relationships is still an open challenge.

\section{Conclusions} \label{sec:conc}
We define multimodal conversational AI and outline the objective function required for its realization. Solving this objective requires multimodal representation and fusion, translation, and alignment. We survey existing datasets and state-of-the-art methods for each sub-task. 
We identify simplifying assumptions made by existing research preventing the realization of multimodal conversational AI. Finally, we outline the collection of a suitable dataset and an approach that utilizes multimodal co-learning as future steps. 

\bibliography{anthology,custom}
\bibliographystyle{acl_natbib}

\clearpage  
\appendix

\section{Conversations with \textsc{TransResNet}}
\label{sec:appendix_transresnet}

\begin{minipage}{1.0\textwidth}
\centering
\begin{adjustbox}{width=\textwidth,center}
\begin{tabu}{ccl}
    \toprule
    Image & Personality & Conversations\\
    \midrule
    \multirow{5}[5]{*}[0.5mm]{\includegraphics[width=3cm, height=2cm]{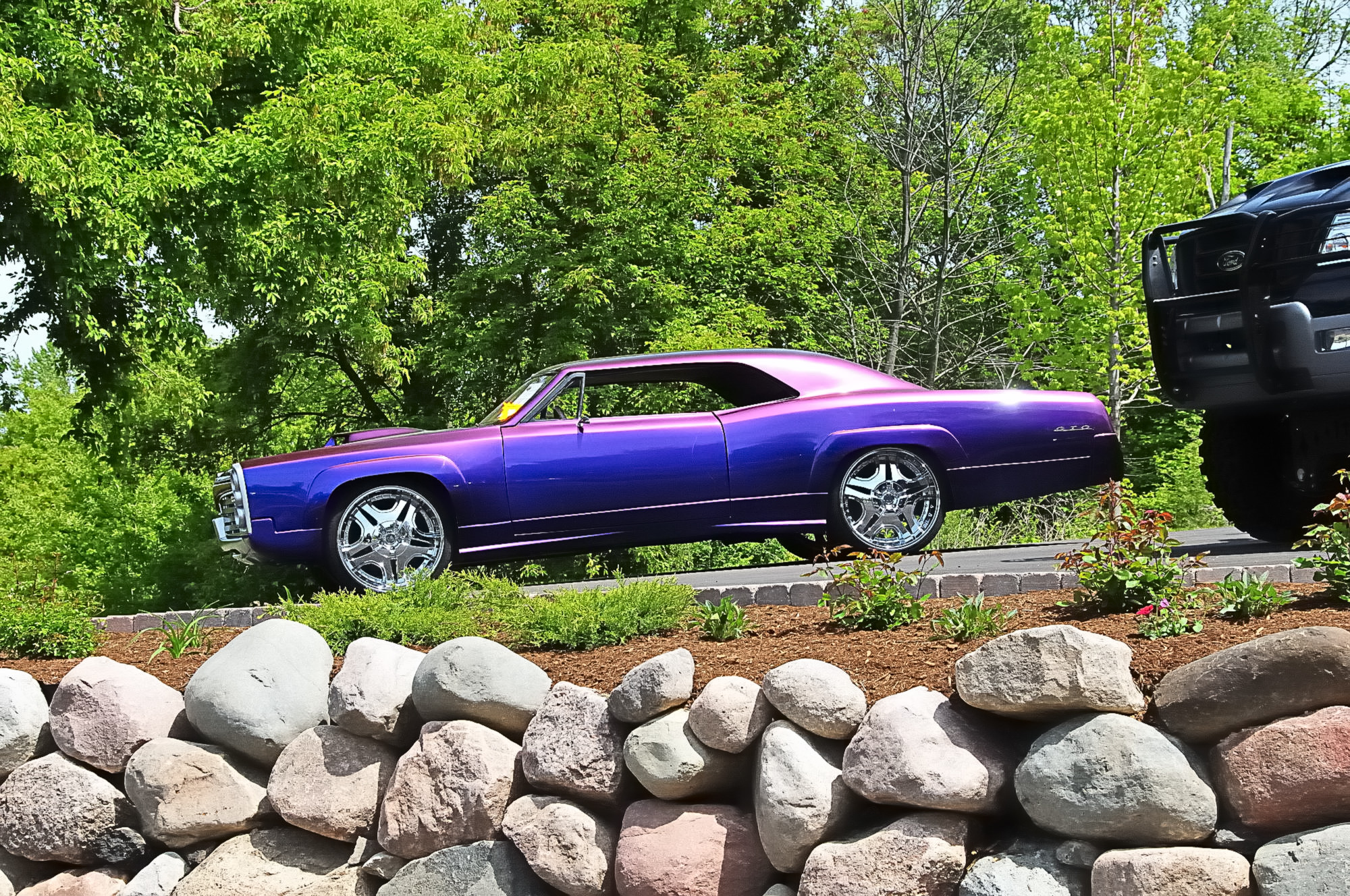}} & Glamorous & The star of the show is the man and his fabulous car \\
    &   [-]     & I don't see a man\\ 
    & Glamorous &  That tends to be the case with men that buy flashy cars and BIG houses! \\
    &  [-]      & I would like to drive that car\\
    & Glamorous & That tends to be the case with men that buy flashy cars and BIG houses!  \\  
    & [-] & Would you drive that car? \\
    & Glamorous & And I would pay him to do that in my backyard as well \\
    \midrule
    \multirow{5}[5]{*}[0.5mm]{\includegraphics[width=3cm, height=2cm]{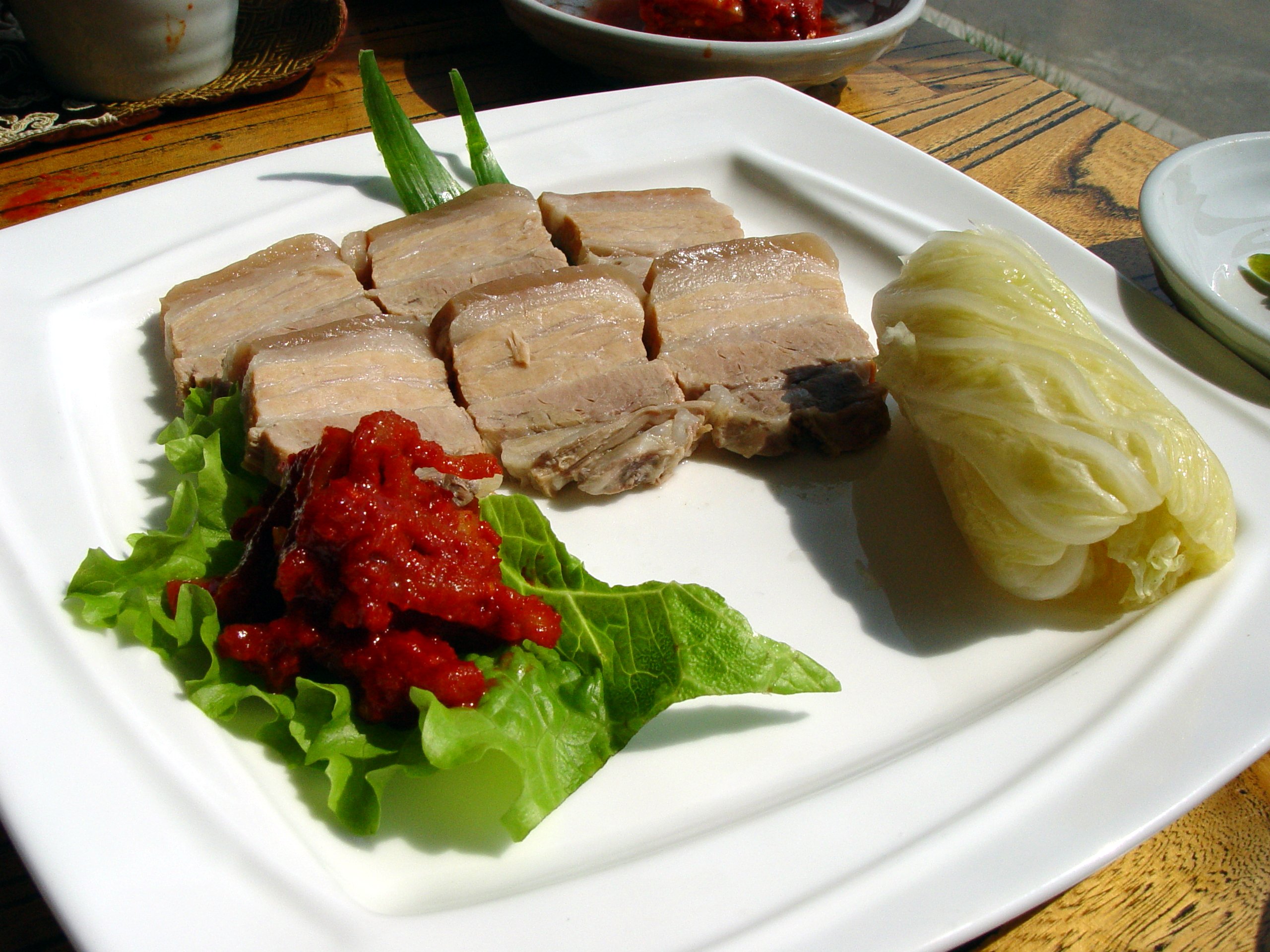}} & [-] & I love going to new restaurants  \\
    & Wise & Well, easy come easy go with your food then, raw is dangerous but it's your life \bigstrut \\
    & [-] & Do you like spicy food? \\
    & Wise & It is most likely too spicy for you \bigstrut  \\
    \midrule
    \multirow{5}[6]{*}[0.5mm]{\includegraphics[width=3cm]{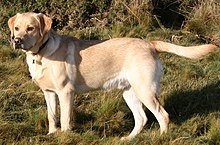}} & [-] & This is Lucy, my labrador  \\
    &   Businesslike     & I should start a dog walking business\\ 
    & [-] &  That's a good idea!  \\
    &  Suave      & He would make a lovely companion for me.\\
    & [-] & Do you have any pets?  \\  
    & Considerate & I think I will. Dogs are a nice animal for the regular person to have\\  
    
    \midrule
    \multirow{5}[6]{*}[0.5mm]{\includegraphics[width=3cm, height=2cm]{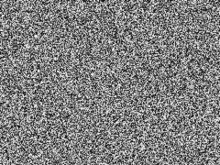}} & Caring & I hope you're injured too. If not, I can help you with that \\
    &   [-]     & I'm not injured\\ 
    & Extreme &  At least if you crash you'll beat the paramedics to the scene by at least 20 minutes! \\
    &  [-]      & Haha that's a nice joke\\
    & Silly & I would die laughing if you did fall\\  
    
    \midrule
    \multirow{12}[6]{*}[0.5mm]{\includegraphics[width=3cm, height=2cm]{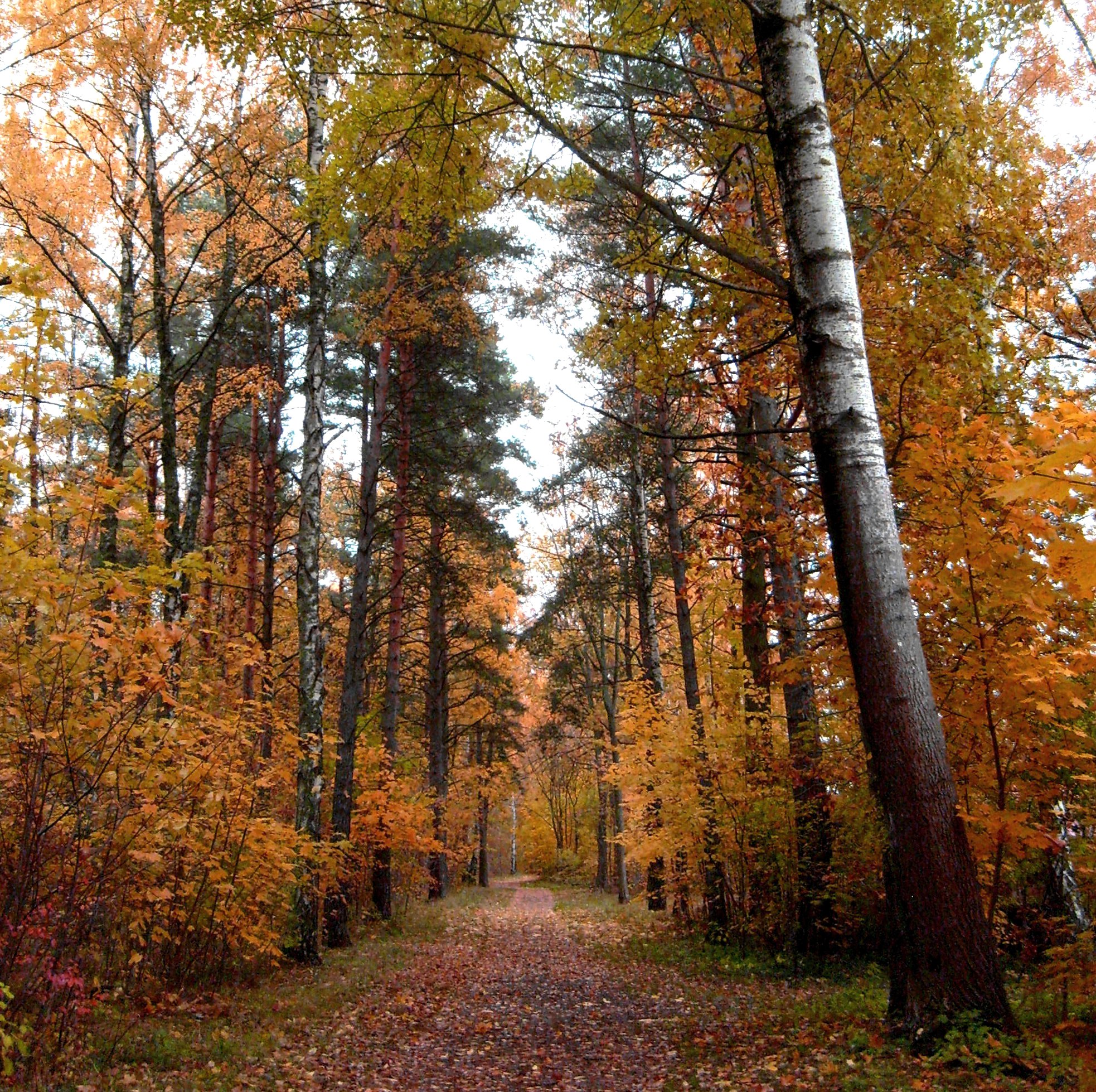}} & [-] & That doesn't look great \\
    &   Angry     & It makes me mad. Why did they have to hang it before the paint was dry?\\
    & [-] & It's not a painting \\
    & Angry & Oh come on, you're so negative! How is that NOT artistic?! \\
    & [-] & I didn't say it wasn't artistic \\
    & Angry & Oh come on, you're so negative! How is that NOT artistic?! \\
    & [-] & I'm not being negative \\
    & Angry & Oh come on, you're so negative! How is that NOT artistic?! \\
    & [-] & Why are you repeating yourself? \\
    & Angry & Because you have ruined the artistic integrity of the image. If you had the talent that I have,\\
    & & you would understand this. \\
    & [-] & I agree with you \\
    & Angry & Why are people so judgemental about art? It's infuriating! \\
    
    \bottomrule

\end{tabu}
\end{adjustbox}
\captionsetup{labelformat=empty}

\captionof{table}{Conversations from \textsc{TransResNet}. Images are chosen from YFCC100M using the browser by \citet{kalkowski_real-time_2015} and the internet, and personalities are chosen at random. }
\label{tab:transresnet}
\text{\footnotesize{[-] indicates lines spoken by the author.}}
\end{minipage}

\end{document}